\documentclass[letterpaper]{article}
\usepackage{aaai25}
\usepackage{times}
\usepackage{helvet}
\usepackage{courier}
\usepackage[hyphens]{url}
\usepackage{graphicx}
\usepackage{natbib} 
\usepackage{caption} 
\usepackage{booktabs}
\usepackage{verbatimbox}
\usepackage{multirow}
\usepackage[table]{xcolor}
\usepackage{algorithm}
\usepackage{algorithmic}
\usepackage{newfloat}
\usepackage{listings}
\DeclareCaptionStyle{ruled}{labelfont=normalfont,labelsep=colon,strut=off}
\floatstyle{ruled}
\newfloat{listing}{tb}{lst}{}
\floatname{listing}{Listing}

\title{Co-Producing AI: Toward an Augmented, Participatory Lifecycle}

\author{
  Rashid Mushkani\textsuperscript{\rm 1,\rm 2}\thanks{Corresponding author}\,,
  Hugo Berard\textsuperscript{\rm 1},
  Toumadher Ammar\textsuperscript{\rm 1},
  Cassandre Chatonnier\textsuperscript{\rm 1},
  Shin Koseki\textsuperscript{\rm 1,\rm 2}
}
\affiliations{
  \textsuperscript{\rm 1}Université de Montréal\\
  \textsuperscript{\rm 2}Mila–Quebec AI Institute\\
  \{rashid.ahmad.mushkani, hugo.berard, toumadher.ammar, cassandre.chatonnier, shin.koseki\}@umontreal.ca
}

\begin{document}

\maketitle

\begin{abstract}
Despite efforts to mitigate the inherent risks and biases of artificial intelligence (AI) algorithms, these algorithms can disproportionately impact culturally marginalized groups. A range of approaches has been proposed to address or reduce these risks, including the development of ethical guidelines and principles for responsible AI, as well as technical solutions that promote algorithmic fairness. Drawing on design justice, expansive learning theory, and recent empirical work on participatory AI, we argue that mitigating these harms requires a fundamental re‑architecture of the AI production pipeline. This re‑design should center co‑production, diversity, equity, inclusion (DEI), and multidisciplinary collaboration. We introduce an augmented AI lifecycle consisting of five interconnected phases: co‑framing, co‑design, co‑implementation, co‑deployment, and co‑maintenance. The lifecycle is informed by four multidisciplinary workshops and grounded in themes of distributed authority and iterative knowledge exchange. Finally, we relate the proposed lifecycle to several leading ethical frameworks and outline key research questions that remain for scaling participatory governance.
\end{abstract}

\section{Introduction}
AI systems are increasingly embedded in decision-critical infrastructures, including law-enforcement identification \cite{Dwivedi2021,Zhao2019}, medical image interpretation, patient-outcome prediction, drug discovery \cite{Yu2018}, and personalized recommendation. The breadth and speed of these deployments create substantial societal leverage.

Empirical evidence indicates that this leverage can intensify existing inequities. Documented harms include predictive-policing bias \cite{Angwin2016}, disparate facial-recognition error rates for people of color \cite{Buolamwini2018}, and discriminatory screening in automated hiring systems \cite{Dastin2022}. These risks are compounded by limited model transparency \cite{Burrell2016}, privacy violations arising from data aggregation \cite{Vliz2021,Nissenbaum2010}, and a broad range of governance challenges cataloged by prior work \cite{Arslan2017,Brayne2017,Galaz2021,Gerdes2018,Golpayegani2022,Koseki2022,Leslie2019,Mohamed2020,Adebayo2022,mushkani2025position,sorensen2024pluralistic}.

Policymakers and professional bodies have responded with high-level guidance. The Montreal Declaration for Responsible AI \cite{montreal2018declaration}, the IEEE Global Initiative on Ethics of Autonomous and Intelligent Systems \cite{ieee2019ethically}, the European Commission’s AI Act \cite{Commission2018}, and the NIST AI Risk Management Framework collectively articulate principles such as autonomy, equity, and accountability \cite{NIST2023}. Yet these instruments remain largely aspirational and center responsibility within expert or organizational hierarchies rather than within affected publics.

Concurrently, technical research has introduced concepts of bias, fairness, and explainability \cite{Arrieta2020,Miller2019}. Numerous fairness criteria, such as statistical parity, equal opportunity, and individual fairness, coexist and sometimes conflict \cite{Bellamy2019,Chouldechova2017,Dwork2012,Hardt2016,Binns2020,Hoffmann2022,Lepri2018}. The multiplicity of metrics, combined with the socio-technical nature of bias \cite{Korobenko2024}, complicates operationalization in practice.

Recent scholarship underscores the cultural contingency of ethical criteria. Studies on Indian art traditions \cite{Divakaran2023} and non-Western honor systems \cite{Wu2023} demonstrate that prevailing frameworks often mirror Western, educated, industrialized, rich, and democratic settings. Cross-cultural surveys reveal divergent alignment preferences \cite{kirk2024prism,mushkani2025livspluralisticalignmentdataset}, and pluralistic-alignment proposals advocate participatory resolution of normative conflict \cite{sorensen2024pluralistic,anthropic2023collective}.

Sector-specific investigations likewise show that ethical challenges vary by context. Domain studies expose distinctive issues in agriculture \cite{Burema2023} and human–computer interaction \cite{Li2022}, while practitioner reflections identify organizational power dynamics that limit the utility of existing ethics toolkits \cite{Wong2023}. These findings suggest that a checklist-based ethics approach is insufficient without continuous stakeholder engagement.

Participatory and co-design traditions offer a partial remedy. Case studies demonstrate improvements in logistics optimization \cite{Berditchevskaia2021}, clinical decision support \cite{Zicari2021}, and public-sector recommender systems \cite{Donia2021}. Nevertheless, most participatory efforts involve brief or single-phase engagements \cite{Gerdes2022,Birhane2022}, and large technology firms still tend to design \emph{for} rather than \emph{with} users \cite{Aizenberg2020}. This gap motivates our research question: \emph{How can citizen participation be integrated throughout the AI lifecycle—balancing process-oriented and outcome-oriented considerations—to produce systems that are both effective and just?}

To address this question, we draw on expansive-learning theory \cite{Engestrm2010} and design-justice principles \cite{CostanzaChock2020}. Building on multidisciplinary workshops \cite{Lember2019,McBride2023}, we propose an augmented AI lifecycle comprising five interconnected phases—co-framing, co-design, co-implementation, co-deployment, and co-maintenance. Each phase positions citizens, domain experts, and technologists as co-producers, thereby operationalizing DEI commitments while fostering iterative knowledge exchange. 

\subsection*{Contributions}
This study makes three key contributions:
\begin{enumerate}
    \item \textbf{Lifecycle proposal:} Synthesizes design-justice and expansive-learning perspectives into a five-phase AI lifecycle that meaningfully includes citizens as co-producers.
    \item \textbf{Ethical mapping:} Maps this lifecycle to prevailing ethical frameworks, highlighting where they converge and diverge from participatory practice.
    \item \textbf{Workshop synthesis:} Distills a shared understanding from four multidisciplinary workshops, of how participatory checkpoints can shape technical and governance choices across the lifecycle.
\end{enumerate}

The remainder of the paper details this lifecycle, relates it to leading ethical frameworks, and outlines future research directions for scalable participatory governance.

\section{Related literature}
\subsection{Lifecycle models across domains}
Product development research identifies three generic phases—beginning-of-life, middle-of-life, and end-of-life—through which artefacts progress \cite{Kiritsis2003}. Early craft models placed all lifecycle tasks in a single agent, as illustrated by the cobbler metaphor, whereas contemporary industrial workflows distribute responsibilities across specialized actors while seeking continuity of data and knowledge \cite{Ameri2005,Ibrahim2008,Terzi2010}. Software engineering codified similar concerns in the software development lifecycle (SDLC), a sequence of planning, analysis, design, implementation, testing, deployment, and maintenance that evolved from the PDCA cycle and agile methods \cite{Deming1986,Beck2001,Mohammed2017,Assal2018}. Machine-learning pipelines inherit these stages but foreground data collection, preprocessing, and algorithm selection. Canonical models present a linear path from problem formulation to deployment, yet in practice, training outcomes routinely prompt returns to prior stages, and pre-trained models introduce additional loops devoted to transfer learning and fine-tuning \cite{Hummer2019,Haakman2021,Sculley2015,Qian2021,Wang2020,Silva2022}. Across domains, effective lifecycle management depends on the unobstructed transfer of artefacts, assumptions, and performance evidence between stages.

\subsection{Participation in the AI lifecycle}
Participatory design emerged in Scandinavian labor contexts and has since influenced a wide range of socio-technical projects \cite{Asaro2000,Birhane2022}. Contemporary methods such as workshops, role-playing, scenario building, prototyping, and other dialogic practices aim to build consensus among diverse stakeholders as they navigate complex and often contested design problems \cite{Flanagan2016,Sloane2022,Fillet2020}.

In the field of AI, participatory approaches have been employed in initiatives such as Collective Constitutional AI \cite{huang2024collective}, the PRISM Alignment Dataset \cite{kirk2024prism}, and MID‑Space \cite{nayak2024midspace}, each of which integrates stakeholder input into the development and evaluation of AI systems. Empirical projects like Project Dorian \cite{berditchevskaia2021participatory} and WeBuildAI \cite{Lee2019} demonstrate the practical value of iterative feedback loops, showing how repeated engagement can shape system outcomes in meaningful ways.

Studies further suggest that involving frontline users in the design and development process can improve task performance and uncover risks that may not be visible to developers alone. This is particularly evident in domains such as logistics, agriculture, public-sector recommender systems, and clinical decision support \cite{Berditchevskaia2021,Zicari2021,Donia2021}.

Despite these advances, most participatory efforts remain limited in scope. Often, stakeholders are consulted during a single phase or brought in for short-term input, while critical aspects such as problem definition, deployment, and post-deployment governance remain in the hands of experts \cite{Gerdes2022,Aizenberg2020,Gerdes2018,Wang2022,Ravanera2021}. This restricted engagement reduces opportunities for meaningful knowledge exchange and limits accountability to the communities most affected by AI systems \cite{Helbing2023,Silva2022}.

\begin{table}[!t]
\centering
\small
\begin{tabular}{@{}p{2.5cm}p{5.3cm}@{}}
\toprule
\textbf{Risk Category} & \textbf{Description} \\
\midrule

Social Responsibility & Ethical duty to consider AI’s societal impacts and promote public benefit. \\
Vulnerable Populations & Risk of exacerbating harm to marginalized or disadvantaged groups. \\
Transparency & Need for clear, understandable AI processes and decisions. \\
Misuse \& Hostile Use & Potential for malicious or unethical applications of AI. \\
Human Rights & Threats to privacy, expression, and freedom from discrimination. \\
Deception \& Manipulation & Use of AI to mislead or covertly influence behavior. \\
Inequalities & Risk of deepening social and economic divides. \\
Workforce Diversity & Importance of inclusive development to avoid narrow perspectives. \\
AGI \& Existential Risk & Long-term concerns over uncontrolled advanced AI. \\
Cultural Sensitivity & Need to respect diverse norms in AI design and use. \\
Trust & Building confidence in AI’s reliability and ethical use. \\
Psychological Impacts & Effects on mental health, identity, and human purpose. \\
Unemployment & Job loss risks from automation-driven displacement. \\
Misinformation & Generation or spread of false information via AI. \\
Economic Growth & AI’s role in boosting innovation and productivity. \\
Explainability & Clarity on how AI makes decisions. \\
Privacy & Safeguarding personal data from misuse or leaks. \\
Safety \& Reliability & Ensuring systems function correctly and safely. \\
Human Control & Maintaining human oversight of AI actions. \\
Accountability & Assigning responsibility for AI outcomes. \\
Consent \& Autonomy & Respecting individuals' control over data and impact. \\
\bottomrule
\end{tabular}
\caption{Risks and ethical considerations in AI development and use}
\label{tab:ai_risks_ethics}
\end{table}

\subsection{Diversity, equity, and inclusion within AI practice}
DEI frameworks hold that individual experiences and identities mediate access to resources and shape technology adoption \cite{Ashley2022,Barton2020,Hond2022}. Reviews of AI deployments reveal persistent bias, limited demographic representation, and uneven privacy impacts, prompting calls for process interventions grounded in DEI principles \cite{CachatRosset2023,Forum2022}. Public, private, and civil-society actors espouse distinct ethical priorities; government and non-profit bodies emphasize broad societal effects and participatory governance, whereas corporate initiatives often concentrate on product-specific risk mitigation \cite{Schiff2021a}. Multidisciplinary collaboration offers one response, aligning technical expertise with normative, legal, and social analysis, yet empirical work documents persistent disciplinary silos and communication barriers \cite{Bikakis2015,Bisconti2023,Dwivedi2021,Beaudouin2020}.

\subsection{Rationale for an augmented AI lifecycle}
The “empirical turn” in technology ethics reframed design as a moral practice, inspiring approaches such as value-sensitive design and ethics-by-design \cite{Umbrello2021,Gerdes2023}. Responsible-AI principles (fairness, accountability, transparency) are now reflected in policy instruments and corporate standards, but operational guidance remains coarse. Competing mathematical definitions of fairness complicate implementation, and high-level frameworks rarely specify mechanisms for citizen participation throughout system development \cite{Barocas2023,Jobin2019,montreal2018declaration,Commission2018,NIST2023}. Comparative analyses show that broad ethical coverage, adaptability across contexts, and iterative governance are necessary but insufficient without practical procedures and documentation that enable stakeholder input \cite{Schiff2021}. Catalogues of harms (discrimination, misinformation, privacy erosion, and others) compile evidence of risks that exceed the remit of traditional lifecycles \cite{Arslan2017,Brayne2017,Galaz2021,Gerdes2018,Golpayegani2022,Koseki2022,mushkani2025negotiative,Mohamed2020,Adebayo2022,gowaikar2024}. Table~\ref{tab:ai_risks_ethics} consolidates these issues. Expansive-learning theory posits that durable solutions emerge when intersecting communities collectively reconstruct their activity systems, while design-justice scholarship centers those most affected by design outcomes \cite{Roth2004,Engestrm2014,Jordan2023}. Together, these perspectives motivate the augmented lifecycle proposed in this work, which embeds co-production, DEI, and iterative knowledge exchange across all phases to address the limitations identified above.

\section{Materials and Methods}
\subsection{Literature Review}
Between October 2023 and May 2024, we conducted a scoping review of academic and gray literature published from January 2013 to May 2024. Our goal was to identify concepts, methods, and empirical findings relevant to ethical and inclusive AI to inform a series of multidisciplinary workshops. We searched Scopus, PubMed, Web of Science, and Google Scholar, covering fields across computer science, the social sciences, and the humanities. We applied the following Boolean expression to titles, abstracts, and keywords:

\texttt{("ethical AI" OR "AI ethics") AND (fairness OR transparency OR accountability OR bias OR "algorithmic fairness" OR "explainable AI" OR "participatory AI" OR "public participation" OR co-production OR co-creation OR "AI ethics guidelines" OR "AI ethics frameworks" OR "human-in-the-loop" OR "responsible innovation" OR "ethical dilemmas" OR "AI for good" OR "value-sensitive design" OR "ethics by design" OR "AI lifecycle" OR "product lifecycle" OR inclusivity OR autonomy OR interpretability)}.

\smallskip
We retained only English-language records. The initial search returned 330 documents. After removing duplicates and screening titles and abstracts for relevance to AI ethics, inclusivity, and lifecycle processes, we narrowed the pool to 147 records. To include a source at the full-text stage, it had to explicitly address (i) ethics and AI, (ii) co-creation or co-production, (iii) lifecycle or process models, and (iv) design-justice perspectives. We also added industrial standards and policy frameworks through targeted searches of organizational repositories. In total, we synthesized 76 sources to prepare the workshop materials.

\subsection{Workshop Design}
Between January and May 2024, we conducted four three-hour workshops in Montréal, Canada. Recruitment followed a purposive sampling strategy designed to ensure disciplinary breadth and institutional diversity. Across the four sessions, we enrolled twenty participants (5–9 per workshop). Participants were affiliated with diverse organizations, including Mila–Quebec AI Institute, Université de Montréal, the Institute for Data Valorization (IVADO), \textit{Institut national de la recherche scientifique} (INRS), and the International Observatory on the Societal Impacts of AI and Digital Technology (OBVIA). Disciplinary backgrounds spanned computer science, social science, law, philosophy, and diversity–equity–inclusion practice. Nine participants identified primarily as researchers, six as industry practitioners, and five as civil-society advocates.

Each workshop began with a summary of key findings from the literature review, followed by a moderated discussion based on three core questions:

\begin{enumerate}
\item How can the ethical challenges identified in the literature be addressed in the design and use of AI systems?
\item Which methods have demonstrated efficacy in producing ethical AI?
\item Which methodological scenarios can guide future ethical AI development?
\end{enumerate}

We recorded audio and took detailed notes, then transcribed and anonymized the data for analysis.

\subsection{Data Analysis}
We used an inductive–deductive approach to code the transcripts thematically. We began with codes informed by expansive-learning and design-justice theory, then added new categories as they emerged until we reached saturation. We resolved coding disagreements through peer debriefing. Once we finalized the codebook, we applied it across all workshop data. The resulting themes form the basis of the augmented AI lifecycle described later in the paper.

\subsection{Theoretical Orientation}
Our analysis draws on expansive-learning and activity theory, which treat learning as a collective transformation of activity systems and boundary crossing between communities \cite{Engestrm2010,Engestrm2014}. We also apply design-justice principles, which center decision-making authority with those most affected by technological outcomes \cite{CostanzaChock2020,Jordan2023}. By combining these frameworks, we treat co-production as a normative requirement rather than an optional supplement \cite{Aizenberg2020,Helbing2023}.

\subsection{Synthesis Procedure}
We triangulated insights from the literature review and workshop data to develop an augmented AI lifecycle with five stages: co-framing, co-design, co-implementation, co-deployment, and co-maintenance. We iterated on this model with workshop participants to ensure it reflected both their contributions and the study’s theoretical foundations.

\section{Results}\label{sec:results}
Thematic analysis of the four workshops produced four interdependent themes that shaped the design of the augmented AI lifecycle. Each theme is grounded in direct participant testimony and reflects recurring patterns observed across sessions.

\paragraph{Distributed authority.}  
Participants consistently argued that decision rights should reside with the communities that will bear the consequences of an AI system. One attendee stated, “Decision-making power has to move closer to the communities who will live with the outcomes; otherwise the system will reproduce existing hierarchies’’ [Workshop 2, P5]. This view aligns with design-justice scholarship and informed the decision to embed community veto rights and shared governance checkpoints throughout the lifecycle.

\paragraph{Iterative knowledge exchange.}  
The workshops highlighted the importance of repeated, dialogic learning. A participant explained that lay contributors acquire AI literacy “through participating in such learning circles where people from different background show up’’ [Workshop 3, P2]. Consequently, the lifecycle specifies cyclic feedback mechanisms—such as community review sessions and shared artifact repositories—that maintain context as teams and project phases evolve.

\paragraph{Contextual privacy.}  
Privacy practices were viewed as inseparable from local social norms. As noted by one participant, “The privacy solution must match local norms; differential privacy alone does not answer the cultural questions’’ [Workshop 1, P4]. This observation supports a layered privacy strategy calibrated to cultural expectations and data sensitivity, extending contextual-integrity arguments \cite{Nissenbaum2010}.

\paragraph{Resource constraints.}  
Sustained engagement was deemed feasible only when budgets addressed tangible costs borne by community members. One participant observed, “We can sustain monthly check-ins only if the budget covers childcare and travel for community members’’ [Workshop 4, P7]. To ensure meaningful involvement rather than tokenistic participation, the engagement lifecycle must include dedicated funding for logistical support and clearly defined role charters \cite{Sloane2022}.

These empirically grounded themes directly informed the tasks, artifacts, and checkpoints detailed in Section~\ref{sec:augmented_lifecycle}.

\section{The Augmented AI Lifecycle}
\label{sec:augmented_lifecycle}

\begin{figure*}[t]
  \centering
  \includegraphics[width=\textwidth]{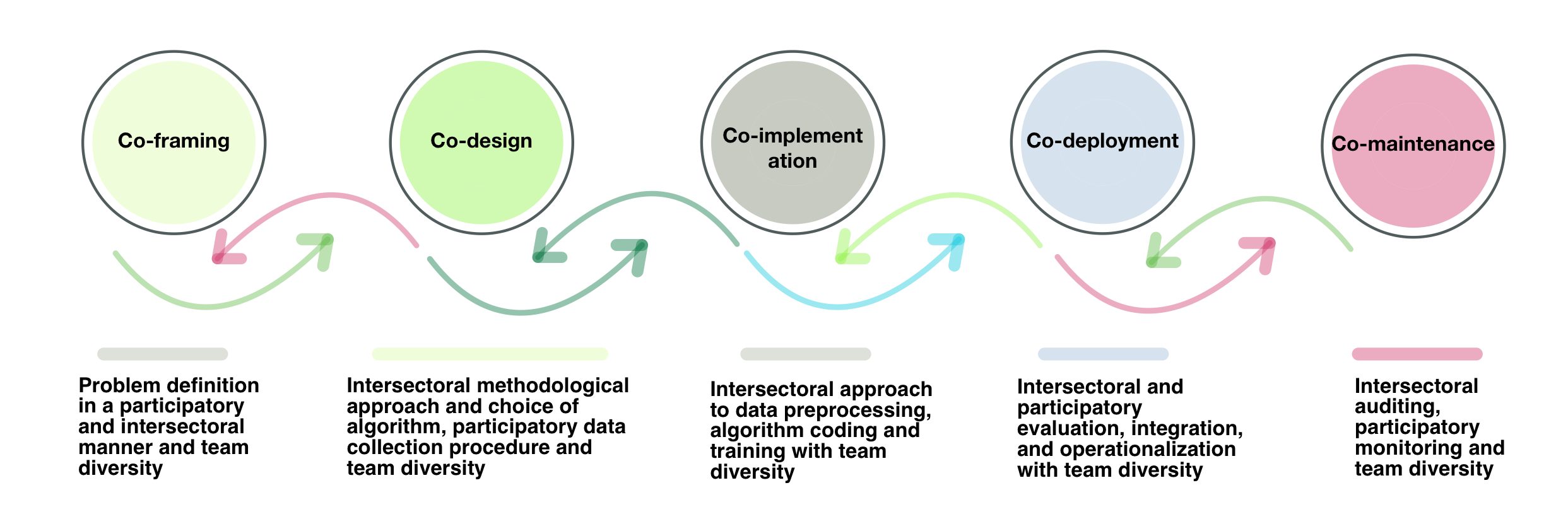}
  \caption{The augmented AI lifecycle. Five co-production phases are linked by continuous knowledge exchange and shared accountability.}
  \label{fig:augmented_ai_lifecycle}
\end{figure*}

\begin{figure*}[t]
  \centering
  \includegraphics[width=\textwidth]{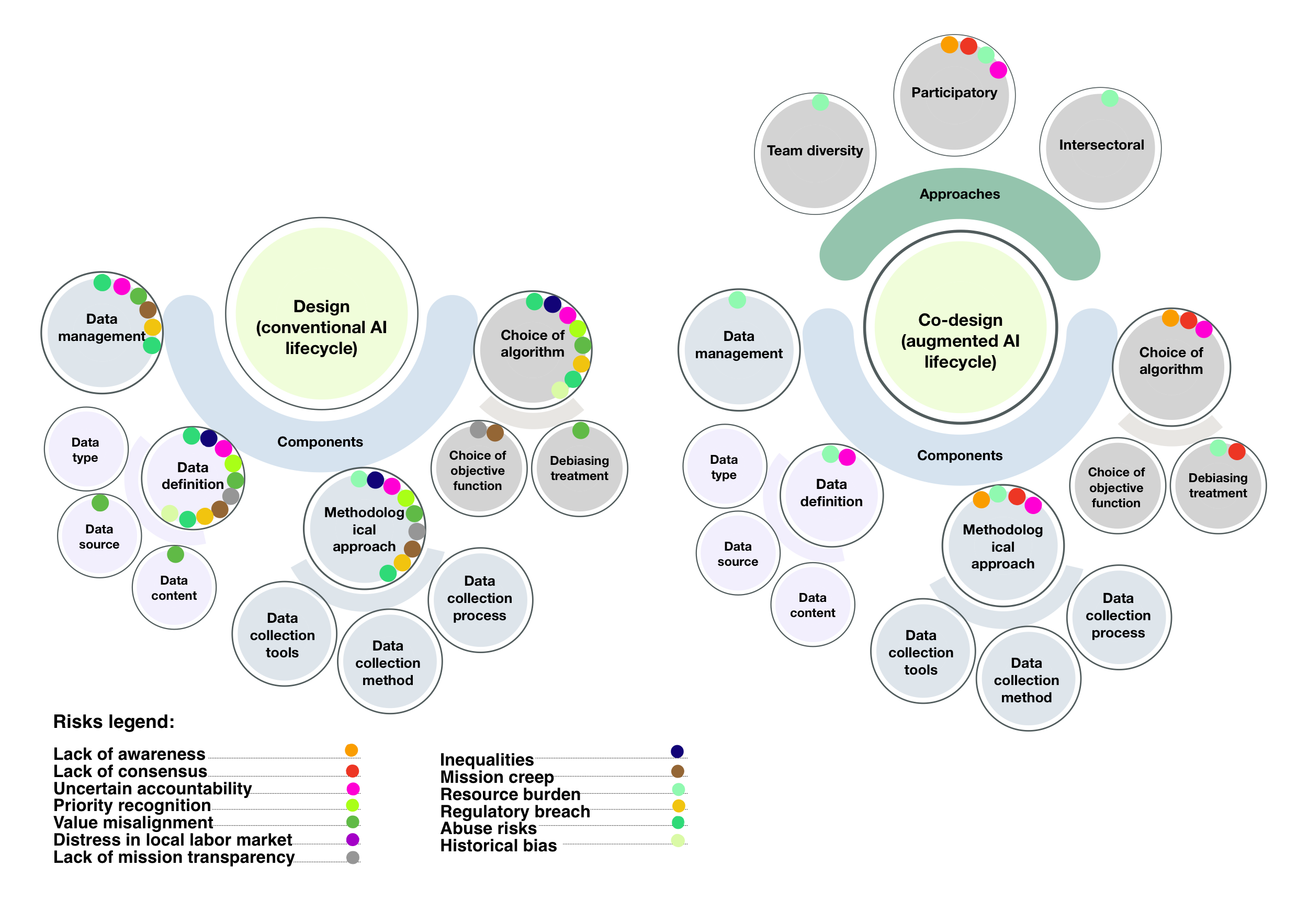}
  \caption{Phase-specific risks in a conventional AI lifecycle and corresponding mitigation via the co-design phase.}
  \label{fig:risk_lifecycle}
\end{figure*}

Grounded in design-justice principles \cite{CostanzaChock2020}, expansive-learning theory \cite{Engestrm2014}, and DEI scholarship \cite{Hond2022,Barton2020}, the augmented AI lifecycle operationalizes co-production across five interdependent phases: \emph{co-framing}, \emph{co-design}, \emph{co-implementation}, \emph{co-deployment}, and \emph{co-maintenance}. Each phase integrates citizens, domain specialists, and technologists as joint decision-makers, thereby addressing gaps identified in conventional lifecycles (Section Rationale for an augmented AI lifecycle). Figure~\ref{fig:augmented_ai_lifecycle} visualizes the overall process, while Figure~\ref{fig:risk_lifecycle} maps design versus co-design phase-specific risks and mitigation strategies derived from our workshops and prior studies \cite{Silva2022,Mitchell2019}.

\subsection{Co-framing}
\label{subsec:co_framing}

Co-framing establishes a shared problem definition, an initial risk register, and a participation plan. Drawing on Arnstein’s ladder of participation \cite{Arnstein1969} and the boundary-crossing mechanisms of expansive learning \cite{Engestrm2010}, the project team convenes citizens likely to experience the system’s outcomes, alongside domain and DEI experts. Structured workshops, semi-structured interviews, and deliberative forums surface contextual knowledge often absent from expert-centric scoping \cite{Koseki2022,Haakman2021}. Key questions include:

\begin{enumerate}
  \item Which communities and activity systems are affected, and how are impacts geographically distributed?
  \item What comparable systems exist, and what ethical failures have been documented?
  \item Which engagement methods (e.g., open calls, community partnerships) align with project constraints and DEI objectives?
  \item How will citizen perspectives reshape the initial problem statement?
\end{enumerate}

\subsection{Co-design}
\label{subsec:co_design}

In this phase, participants select data sources, model families, and interface concepts aligned with co-framed objectives. Participatory prototyping sessions and scenario walk-throughs translate workshop insights into technical specifications \cite{Fillet2020,Sloane2022}. Comparative risk assessments evaluate trade-offs among pipeline options, clarifying how choices distribute benefits and burdens \cite{Berditchevskaia2021,Zicari2021}. Guiding questions include:

\begin{enumerate}
  \item How do data-collection strategies affect representation and privacy?
  \item Which foundational models, if any, meet performance requirements without compromising interpretability or fairness?
  \item How does the overall pipeline architecture support autonomy and privacy while minimizing environmental impact?
  \item What documentation makes design decisions accessible to non-technical stakeholders?
  \item How will participatory feedback be incorporated when reusing pre-trained components?
\end{enumerate}

\subsection{Co-implementation}
\label{subsec:co_implementation}

This phase includes data acquisition, feature engineering, model training/fine-tuning, and iterative validation. A multidisciplinary team records decisions in version-controlled artifacts and model cards to ensure traceability to co-design deliberations \cite{Mitchell2019}. Citizen partners review intermediate outputs, such as data summaries and error reports, to check conformance with DEI commitments \cite{CachatRosset2023}. Guiding questions include:

\begin{enumerate}
  \item Does the implementation team reflect heterogeneous expertise and lived experience?
  \item How are privacy safeguards (e.g., differential privacy, k-anonymity) applied and communicated \cite{Pawar2018,Vliz2021}?
  \item Which artifacts (code, configuration, audit trails) are required for downstream accountability?
  \item How is knowledge transferred to citizen partners to support informed oversight?
\end{enumerate}

\subsection{Co-deployment}
\label{subsec:co_deployment}

Deployment introduces the system into non-stationary social contexts where latent risks may emerge \cite{Koseki2022}. Citizen representatives conduct user-acceptance testing with domain experts, while a governance charter assigns responsibility for emergent harms. Performance dashboards and recourse mechanisms enable affected individuals to contest outputs \cite{Galaz2021}. Central questions include:

\begin{enumerate}
  \item Have evaluation metrics captured distributional impacts across stakeholder groups?
  \item What safeguards prevent mission creep and unauthorized secondary use?
  \item How long is the model considered valid, and what triggers rollback or redesign?
  \item What transparency measures communicate real-time system behavior to the public?
\end{enumerate}

\subsection{Co-maintenance}
\label{subsec:co_maintenance}

Co-maintenance treats the system as a living artifact subject to concept drift, regulatory change, and evolving norms \cite{Silva2022}. Periodic audits (technical, ethical, and participatory) evaluate alignment with original objectives. Citizen assemblies or standing committees may recommend updates, suspension, or decommissioning \cite{Helbing2023}. Key questions include:

\begin{enumerate}
  \item What cadence and scope govern multidisciplinary audits, and who funds them?
  \item How are updates communicated, and how is consent re-established when functionality changes?
  \item What dispute-resolution processes address tensions between commercial aims and public interest?
  \item How is institutional memory preserved as team membership changes?
\end{enumerate}

\subsection{Cross-cutting Considerations}
Across all phases, successful co-production depends on (i) recruitment practices that meet DEI goals \cite{Barton2020}, (ii) anonymization protocols that uphold contextual integrity \cite{Nissenbaum2010,Vliz2021}, and (iii) facilitation methods that reduce power asymmetries \cite{Turnhout2010,Jordan2023}. Our workshop participants shared a common understanding that participatory practices are feasible within typical resource constraints, provided that roles, incentives, and communication channels are clearly defined from the outset.

\smallskip
Overall, the augmented lifecycle embeds shared accountability and iterative learning throughout AI development, aligning empirical methods with the normative goals of policy frameworks and design-justice scholarship. The following sections assess its applicability across domains and propose avenues for scalable participatory governance.

\section{Discussion}
\label{sec:discussion}
This discussion interprets the findings of the workshops using the conceptual frameworks of design justice, expansive learning, and DEI. These lenses help situate how the proposed lifecycle addresses the methodological and normative commitments surfaced during the study. Rather than suggesting deterministic outcomes, the discussion outlines contingent pathways through which co-production may support negotiated responses to AI-related harms.

\subsection{Engaging Participants Through Design-Justice Recruitment}
Design-justice literature positions affected communities as essential actors in shaping technology. Workshop discussions highlighted the importance of beginning engagement with the identification of three domains: \textit{area of impact}, \textit{area of interest}, and \textit{project constraints}. Participants described this as a prerequisite for identifying appropriate community stakeholders and clarifying expectations among developers. They further emphasized that early transparency regarding project scope and resource constraints can mitigate asymmetries in authority \cite{Helbing2023,mushkani2025position}. Several recruitment practices, such as open calls, partnerships with community organizations, and collaboration with labor unions, were discussed as potentially viable, contingent on alignment with DEI goals and access to compensation mechanisms.

\subsection{Participation Modalities as Expansive-Learning Interventions}
Expansive-learning theory conceptualizes participation as an encounter between intersecting activity systems. In the workshops, participants assessed various engagement formats in relation to these boundary-crossing dynamics. In-person sessions were described as productive for surfacing tacit knowledge, although resource-intensive \cite{Creswell2013}. Online tools expanded geographic reach but were noted to risk excluding participants with limited connectivity \cite{Goggin2022}. Other modalities, such as surveys and deliberative assemblies, were associated with distinct trade-offs concerning inclusivity, cost, and depth of interaction \cite{Goggin2019,Lewis2020,Turnhout2010}. These insights suggest that no single method guarantees equitable participation. Instead, formats may need to be selected or combined based on contextual factors, including institutional capacity and participant access.

\subsection{Sustaining Partnerships Across the Lifecycle}
\begin{figure}[t]
  \centering
  \includegraphics[width=3.5cm]{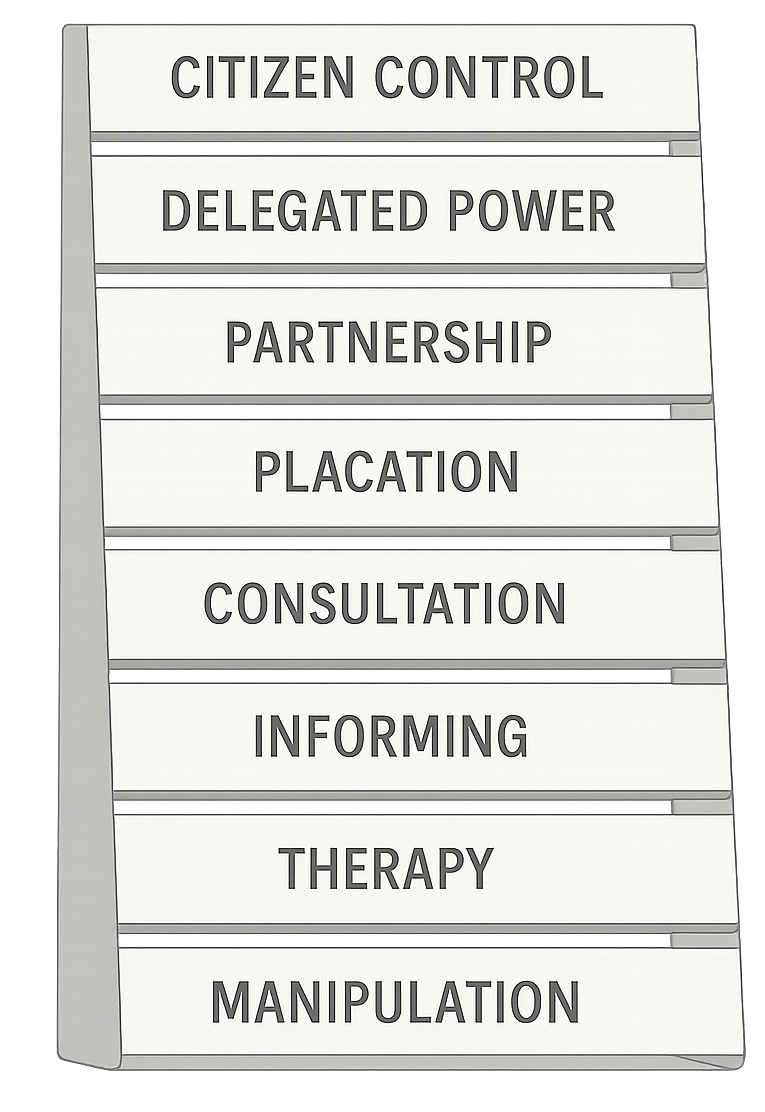}
  \caption{Arnstein’s ladder of citizen participation. A visual framework showing eight levels of citizen involvement, from nonparticipation to full citizen control in decision-making.}
  \label{fig:ladder}
\end{figure}

Long-term legitimacy depends on structures that sustain dialogue beyond the initial design phase. Participants recommended institutionalizing “boundary spaces” such as public innovation labs, recurring town hall meetings, and virtual forums. These venues support ongoing monitoring of concept drift and shifts in social norms, aligning with the principles of the co-maintenance phase. As illustrated in Figure~\ref{fig:ladder}, Arnstein’s ladder places such arrangements at the “partnership” level, where citizen influence extends to shared authority over strategic decisions \cite{Arnstein1969}. Establishing clear charters that define roles, compensation, and audit rights is essential for maintaining stakeholder engagement over multi-year timeframes \cite{Chambers2022}.

\subsection{Multidisciplinary Collaboration and Team Diversity}
Workshop testimony indicated that projects requiring both social and technical competencies often involve coordination across disciplines, including computer science, DEI, social science, and law \cite{Bisconti2023,Hond2022}. Participants described the benefits of this arrangement as situational and contingent. While some reported gains in creativity and reflexivity, others noted the challenge of negotiating divergent epistemologies and terminologies \cite{Sloane2022}. Participants proposed strategies such as rotating facilitation, shared glossaries, and just-in-time expert input to manage friction points \cite{Avellan2020,CachatRosset2023}.

Building consensus can be achieved throughout the discourse and decision-making process by allowing adequate time for all participants to express their opinions—for example, through the use of Indigenous tools like talking sticks or methods such as voting. Moreover, treating each partner equally, valuing their input, and avoiding power imbalances can promote consensus and agreement during the discourse \cite{Lewis2020,Turnhout2010}. Building consensus strengthens the legitimacy and acceptance of outcomes through inclusive dialogue and shared decision-making. It is considered crucial in conflict resolution and community-based planning but requires patience, open-mindedness, and often, neutral facilitation to navigate differing viewpoints \cite{Chambers2022}.

\subsection{Privacy, Anonymity, and Trust}
Sustained collaboration likely requires robust local privacy guarantees. Participants conveyed a shared understanding that a layered approach—beginning with informed consent and progressing through k-anonymity, \textit{l}\,--diversity, \textit{t}\,--closeness, pseudonymization, and differential privacy—can be calibrated to the sensitivity of the data \cite{Pawar2018,Vliz2021,Nissenbaum2010}. Onion-routing techniques protected communication channels during remote sessions \cite{Obaidat2020}. These measures, combined with regular transparency reports, appeared to strengthen trust and reduce attrition among citizen partners. Self-identification questionnaires enabled demographic monitoring without disclosing individual identities, aligning data stewardship practices with DEI objectives \cite{Saunders2018}.

\begin{table*}
\centering
\small
\begin{tabular}{p{4.2cm}p{4.5cm}ccc}
\toprule
\textbf{Title} & \textbf{Principles} & \textbf{DEI Participation} & \textbf{Multidisciplinarity} & \textbf{Team Diversity} \\
\midrule

\multirow{10}{*}{\parbox{4.2cm}{The Montreal Declaration\\for Responsible AI}}
 & Well-being & \cellcolor{gray!25} & \cellcolor{gray!25} & \cellcolor{orange!25} \\
 & Respect for autonomy & \cellcolor{gray!25} & \cellcolor{gray!25} & \cellcolor{gray!25} \\
 & Protection of privacy and intimacy & \cellcolor{gray!25} & \cellcolor{gray!25} & \cellcolor{gray!25} \\
 & Solidarity & \cellcolor{gray!25} & \cellcolor{gray!25} & \cellcolor{gray!25} \\
 & Democratic participation & \cellcolor{orange!25} & \cellcolor{gray!25} & \cellcolor{green!25} \\
 & Equity & \cellcolor{orange!25} & \cellcolor{gray!25} & \cellcolor{green!25} \\
 & Diversity inclusion & \cellcolor{green!25} & \cellcolor{gray!25} & \cellcolor{green!25} \\
 & Caution & \cellcolor{gray!25} & \cellcolor{gray!25} & \cellcolor{gray!25} \\
 & Responsibility & \cellcolor{gray!25} & \cellcolor{gray!25} & \cellcolor{gray!25} \\
 & Sustainable development & \cellcolor{gray!25} & \cellcolor{gray!25} & \cellcolor{gray!25} \\
\midrule

\multirow{6}{*}{\parbox{4.2cm}{Microsoft Responsible\\AI Standard}}
 & Accountability & \cellcolor{gray!25} & \cellcolor{gray!25} & \cellcolor{gray!25} \\
 & Transparency & \cellcolor{gray!25} & \cellcolor{gray!25} & \cellcolor{gray!25} \\
 & Fairness & \cellcolor{green!25} & \cellcolor{green!25} & \cellcolor{green!25} \\
 & Reliability \& safety & \cellcolor{gray!25} & \cellcolor{gray!25} & \cellcolor{gray!25} \\
 & Privacy \& security & \cellcolor{gray!25} & \cellcolor{gray!25} & \cellcolor{gray!25} \\
 & Inclusiveness & \cellcolor{green!25} & \cellcolor{green!25} & \cellcolor{green!25} \\
\midrule

\multirow{8}{*}{IEEE Global Initiative}
 & Human rights & \cellcolor{gray!25} & \cellcolor{gray!25} & \cellcolor{gray!25} \\
 & Well-being & \cellcolor{green!25} & \cellcolor{green!25} & \cellcolor{green!25} \\
 & Data agency & \cellcolor{green!25} & \cellcolor{green!25} & \cellcolor{green!25} \\
 & Effectiveness & \cellcolor{green!25} & \cellcolor{green!25} & \cellcolor{green!25} \\
 & Transparency & \cellcolor{gray!25} & \cellcolor{gray!25} & \cellcolor{gray!25} \\
 & Accountability & \cellcolor{gray!25} & \cellcolor{gray!25} & \cellcolor{gray!25} \\
 & Awareness of misuse & \cellcolor{gray!25} & \cellcolor{gray!25} & \cellcolor{gray!25} \\
 & Competence & \cellcolor{green!25} & \cellcolor{green!25} & \cellcolor{green!25} \\
\midrule

\multirow{7}{*}{\parbox{4.2cm}{AI Risk Management\\Framework — NIST}}
 & Valid and reliable & \cellcolor{green!25} & \cellcolor{green!25} & \cellcolor{green!25} \\
 & Safe & \cellcolor{green!25} & \cellcolor{green!25} & \cellcolor{green!25} \\
 & Secure and resilient & \cellcolor{green!25} & \cellcolor{green!25} & \cellcolor{green!25} \\
 & Accountable and transparent & \cellcolor{gray!25} & \cellcolor{gray!25} & \cellcolor{gray!25} \\
 & Explainable and interpretable & \cellcolor{green!25} & \cellcolor{green!25} & \cellcolor{green!25} \\
 & Privacy-enhanced & \cellcolor{green!25} & \cellcolor{green!25} & \cellcolor{green!25} \\
 & Fair with harmful bias managed & \cellcolor{green!25} & \cellcolor{green!25} & \cellcolor{green!25} \\
\midrule

\multirow{7}{*}{\parbox{4.2cm}{Ethics Guidelines for\\Trustworthy AI — EC}}
 & Human agency and oversight & \cellcolor{gray!25} & \cellcolor{gray!25} & \cellcolor{gray!25} \\
 & Technical robustness and safety & \cellcolor{green!25} & \cellcolor{green!25} & \cellcolor{green!25} \\
 & Privacy and data governance & \cellcolor{green!25} & \cellcolor{green!25} & \cellcolor{green!25} \\
 & Transparency & \cellcolor{gray!25} & \cellcolor{gray!25} & \cellcolor{gray!25} \\
 & Diversity, discrimination, \& fairness & \cellcolor{green!25} & \cellcolor{green!25} & \cellcolor{green!25} \\
 & Environmental \& societal well-being & \cellcolor{green!25} & \cellcolor{green!25} & \cellcolor{green!25} \\
 & Accountability & \cellcolor{gray!25} & \cellcolor{gray!25} & \cellcolor{gray!25} \\
\bottomrule
\end{tabular}
\caption{Alignment of selected AI ethics guidelines with dimensions of co-production in the AI lifecycle. Shaded cells indicate levels of alignment: light gray denotes high correspondence, while light green indicates areas needing further research. This comparison highlights opportunities for future inquiry and practice.}
\label{tab:guideline_alignment}
\end{table*}

\begin{table*}
\centering
\small
\begin{tabular}{p{4cm}p{12cm}}
\toprule
\textbf{Impact Category} & \textbf{Description} \\
\midrule
Enhanced AI Reliability & By incorporating diverse perspectives from the co-framing to the co-maintenance phase, AI systems are likely to be more robust and reliable. \\
Minimized Biases & Active involvement of citizens, especially from marginalized communities, can help identify and rectify inherent biases in AI algorithms. \\
Contextual Solutions & AI systems co-produced with citizens are likely to be contextually relevant, addressing real-world challenges effectively. \\
Empowered Communities & The approach not only democratizes the AI development process but also empowers communities by giving them a voice in technology creation. \\
\bottomrule
\end{tabular}
\caption{Impact of the augmented AI lifecycle}
\label{tab:augmented_ai_impact}
\end{table*}

\section{Limitations and Future Work}
\label{sec:limitations_futurework}

This study is subject to three main limitations that circumscribe the evidentiary strength and external validity of its findings.  
\textbf{First}, data collection relied on four workshops held in Montréal, Canada, with twenty self-selected participants drawn from research, industry, and civil-society organizations. Lay citizens and communities situated outside the local AI ecosystem were not included. The resulting sample is small, geographically concentrated, and professionally homogeneous, which restricts transferability to less-resourced or culturally distinct settings and increases the likelihood that a limited number of voices shaped the consensus \cite{Sloane2022,Bisconti2023}. Social-desirability pressures may have further muted dissent despite anonymized transcription.

\textbf{Second}, the scoping review covered English-language literature. While we strived to include a broad range of sources, in-depth discussion of some literature did not occur. Due to the format of the workshops, we couldn't engage the literature systematically. No formal quality appraisal of individual studies was performed; therefore, the reliability of specific contributions varies across sources.

\textbf{Third}, the augmented lifecycle remains a conceptual framework whose practical feasibility, cost profile, and performance implications have not been empirically evaluated. The workshops provided preliminary face validity, but they do not establish that co-production systematically reduces algorithmic harms or enhances technical metrics. Potential conflicts between community veto rights and organizational accountability mechanisms identified in prior work \cite{Helbing2023,Silva2022} were not examined in operational environments.

Future research should address these constraints by (i) conducting longitudinal field trials that involve citizen cohorts—including historically marginalized groups—in multiple jurisdictions, (ii) extending the evidence base to incorporate non-English and gray sources, and (iii) collecting quantitative and qualitative data on fairness, privacy, cost, and project efficiency when the lifecycle is applied in production settings. Comparative studies across regulatory regimes will clarify how the framework interacts with emerging legislation such as the EU AI Act \cite{Commission2018}. Systematic cost–benefit analyses are also required to determine whether the additional governance overhead yields proportional societal value. While conceptual, this lifecycle offers a structured response to a widely acknowledged gap in operational AI ethics.

While conceptual, the proposed lifecycle offers a structured response to a widely acknowledged gap in operational AI ethics—namely, the absence of participatory, accountable processes that center the voices of those most affected by AI systems.

\section{Risks and Challenges}
\label{sec:risks}
Analysis of the workshops suggests that risks can emerge at various stages of AI development and are not evenly distributed across social groups. Participants, drawing on design-justice principles, emphasized that communities with limited institutional power often experience disproportionate exposure to these risks and should be included in risk-mitigation processes from the outset. Expansive-learning theory provides a framework through which these processes may be understood as opportunities for collective reframing of the AI activity system. Commitments to diversity, equity, and inclusion (DEI) inform which stakeholders are necessary for participation in these reframing processes.

Existing guidelines and empirical studies identify recurring patterns such as feedback loops that reinforce structural inequities, system opacity that limits contestability, and mission drift that expands AI use beyond its intended scope \cite{Koseki2022}. Leslie categorizes the resulting harms into six types: bias and discrimination; non-transparency; social isolation; denial of autonomy, recourse, and rights; unreliable or unsafe outcomes; and privacy invasion \cite{Leslie2019}. Mohamed et al. contextualize these harms within broader post-colonial power asymmetries \cite{Mohamed2020}. The workshop discussions were consistent with these classifications and further identified context-specific concerns, including labor displacement in localized economies and the implications for Indigenous data sovereignty.

The augmented AI lifecycle proposes mechanisms for addressing such risks by incorporating co-production checkpoints at each developmental phase. In earlier stages, inclusion of citizen and domain experts facilitated the surfacing of latent value conflicts. Iterative knowledge exchange supported the identification and modification of proxy variables that were associated with socioeconomic status. During co-implementation, community feedback on misclassification logs led to additional data collection intended to address representational imbalances. These instances demonstrate how mechanisms associated with expansive learning—such as boundary crossing, negotiation of meaning, and iterative model adjustment—can serve as instruments for operationalizing design-justice principles.

The co-production approach introduces several challenges. Diverse perspectives can complicate consensus formation and increase the resource demands of a project \cite{Sloane2022,Varanasi2023}. Meeting DEI-related participation goals often requires sustained outreach and adequate compensation, which may be under-prioritized in institutional planning. Multidisciplinary collaboration may be affected by epistemic divergence and differences in disciplinary language, leading to elevated coordination costs \cite{Bisconti2023,Bikakis2015}. Ambiguities in role definitions can give rise to participation washing, where engagement is formal rather than substantive \cite{Sloane2022}. Addressing these issues may require governance frameworks, including decision-making charters, facilitation tools to support cross-disciplinary communication, and periodic audits to assess the outcomes of participatory interventions \cite{Birhane2022}.

Table~\ref{tab:augmented_ai_impact} presents a summary of workshop participants’ observations regarding the impacts and trade-offs associated with the augmented lifecycle. This framework provides one possible approach for managing tensions between technical performance goals and social accountability. Table~\ref{tab:guideline_alignment} outlines its alignment with established ethical guidelines.

\section{Conclusion}
\label{sec:conclusion}

This study integrates design-justice scholarship, expansive-learning theory, and DEI research into an augmented AI lifecycle that operationalizes co-production across five interdependent phases. A scoping review and four multidisciplinary workshops provided empirical grounding for the model and demonstrated concrete mechanisms (co-framing, co-design, co-implementation, co-deployment, and co-maintenance) that redistribute decision-making authority toward affected publics.

The lifecycle remains conceptual until validated in applied settings. Future work should therefore implement the framework in domains such as health, finance, and public administration, measuring its effects on system performance, participant satisfaction, and long-term accountability. Comparative studies are required to evaluate scalability across organizational contexts and to analyze cost–benefit trade-offs relative to conventional pipelines. Further investigation of privacy, labor, and innovation outcomes will refine governance recommendations and inform alignment with evolving regulatory standards.

By situating technical choices within participatory structures that reflect diverse expertise and lived experience, the augmented AI lifecycle offers a practical pathway toward AI systems that are both contextually responsive and socially just.

\bibliography{aaai25}

\end{document}